\documentclass[sn-mathphys-num]{sn-jnl}

\usepackage{graphicx}%
\usepackage{multirow}%
\usepackage{amsmath,amssymb,amsfonts}%
\usepackage{amsthm}%
\usepackage{mathrsfs}%
\usepackage[title]{appendix}%
\usepackage{xcolor}%
\usepackage{textcomp}%
\usepackage{manyfoot}%
\usepackage{booktabs}%
\usepackage{algorithm}%
\usepackage{algorithmicx}%
\usepackage{algpseudocode}%
\usepackage{listings}%

\theoremstyle{thmstyleone}%
\theoremstyle{thmstyletwo}%

\theoremstyle{thmstylethree}%

\begin{document}

\title[Article Title]{Integrating Functionalities To A System Via Autoencoder Hippocampus Network}


\author*[1]{\fnm{Siwei} \sur{Luo}}\email{luosiwei@jxnu.edu.cn}



\affil*[1]{\orgdiv{Software School}, \orgname{Jiangxi Normal University}, \orgaddress{\street{No.99 Ziyang Avenue, High Technology Development Zone}, \city{Nanchang}, \postcode{330022}, \state{Jiangxi}, \country{China}}}




\abstract{Integrating multiple functionalities into a system poses a fascinating challenge to the field of deep learning. While the precise mechanisms by which the brain encodes and decodes information, and learns diverse skills, remain elusive, memorization undoubtedly plays a pivotal role in this process. In this article, we delve into the implementation and application of an autoencoder-inspired hippocampus network in a multi-functional system. We propose an autoencoder-based memorization method for policy function's parameters. Specifically, the encoder of the autoencoder maps policy function's parameters to a skill vector, while the decoder retrieves the parameters via this skill vector. The policy function is dynamically adjusted tailored to corresponding tasks. Henceforth, a skill vectors graph neural network is employed to represent the homeomorphic topological structure of subtasks and manage subtasks execution.}

\keywords{Dynamical hierarchical reinforcement learning; Autoencoder hippocampus network; Graph neural network; Tasks decomposition}



\maketitle

\section{Introduction}\label{sec1}

Recently, deep learning models, designed and optimized to excel at a particular type of problem, are task-specific. Multi-functionality capability is a fascinating topic. Memorization play a important role in decision-making process. 

The crucial structure hippocampus locates in the medial temporal lobe of the brain, specifically within the limbic system\cite{Eichenbaum}. Named for its resemblance to the shape of a seahorse, the hippocampus plays a vital role in various cognitive functions, particularly memory. The hippocampus is composed of several distinct regions, including the dentate gyrus, the cornu ammonis fields, and the subiculum. These regions work together to facilitate the encoding, consolidation, and retrieval of memories. The hippocampus is integral to this process, helping to encode and store these memories for later retrieval. In addition to its role in declarative memory, the hippocampus also plays a significant role in spatial navigation and orientation. It is involved in the formation of cognitive maps, which help us navigate our environment and understand spatial relationships. The hippocampus is highly interconnected with other brain regions, including the cortex, amygdala, and thalamus. These connections allow it to integrate information from various sources and coordinate cognitive processes. The hippocampus is a vital brain structure that plays a crucial role in memory, spatial navigation, and orientation.

The neuroimaging technique Functional magnetic resonance imaging (fMRI) measures changes in blood flow and oxygen consumption in the brain associated with neuronal activity\cite{Heeger}. This technology utilizes magnetic fields and radio waves to produce images of the brain that highlight regions of activation during various cognitive tasks or in response to external stimuli. When neurons are active, they consume more oxygen and nutrients, leading to increased blood flow in the local region to meet the metabolic demands. Specifically, fMRI relies on the blood oxygen level-dependent (BOLD) contrast. The BOLD signal reflects the relative concentration of oxygenated and deoxygenated hemoglobin in the blood\cite{Logothetis}. When neurons are active, the local blood flow increases, leading to an increased concentration of oxygenated hemoglobin (relative to deoxygenated hemoglobin) in the blood. This change in the oxygenation level of the blood results in a change in the magnetic properties of the blood, which can be detected by the fMRI scanner. During fMRI, the scanner measures the changes in the BOLD signal over time as a subject performs a cognitive task or is exposed to an external stimulus. These changes in the BOLD signal are then used to create images that highlight the regions of the brain that are activated during the task or stimulus. These images provide insights into the neural processes underlying cognitive functions.

Dynamical Hierarchical Reinforcement Learning (DHRL) is an active and advanced field within reinforcement learning that incorporates hierarchical structures to tackle complex and dynamic environments\cite{Haarnoja,Stulp}. The essential idea of DHRL is decompose task into subtasks, solving the task in divide-and-conquer manner, aiming to reduce the complexity of task and enhance learning efficiency that subject to long-term dependencies and sparse reward signal in reinforcement learning. Hierarchical task graph decompose task into subtasks, every vertice in graph represents a subtask and corresponding edge between subtasks indicates calling relationship. The dependencies and priorities among subtasks are concerned by DHRL and guarantee the accomplishment of a task. Maxq method decomposes the value function of a complex task into the sum of value functions of several subtasks. By doing so, it significantly reduces the number of state-action pairs to be considered, thereby improving learning efficiency\cite{Dietterich}. FeUdal Networks\cite{Dayan} adopt a manager-worker architecture, where high-level managers set goals and low-level workers execute the specific actions to achieve these goals. This architecture enables the model to effectively handle long-term dependencies.

Although DHRL demonstrates advantages and potential in the self-adaptive learning process, the practice suggests that we shall better separate memorization from task management. The very observation is: first, a specific region in the brain is in charge of memory and plays a crucial role in this procedure of coordination and execution; second, the brain doesn't change the structure during the procedure but consume different amount of resources. To mimic this process, this paper introduces a design that integrate functionalities to a single system architecture via autoencoder memorization mechanism and tasks graph management. The main contributions of this article are summarized as below: 
1) memorization mechanism is implemented via Autoencoder Hippocampus network;
2) motivation mechanism is implemented and embedded into the system via graph neural network.

\section{Supervised learning based on classical control theory}\label{sec2}

Reinforcement learning techniques enables agent to learn optimal behavior strategies through interactions with environment\cite{Kaelbling}. It can be used to train agent in a self-supervised manner, without the need for explicit human supervision or labeling observation-action pair datasets. This reduces the amount of human effort required for training and makes the process more scalable. Significant progress has been made in terms of algorithmic advancements, simulation environments, and computing power. Advantage Actor-Critic(A2C)\cite{Chu}, Asynchronous Advantage Actor-Critic(A3C)\cite{Du}, and Proximal Policy Optimization(PPO)\cite{Mazyavkina} are reinforcement learning algorithms that utilize different techniques to learn optimal policies for sequential decision-making problems. A2C and A3C utilize the Actor-Critic framework with an advantage function, while PPO focuses on achieving stable and efficient policy updates through a clipped surrogate objective function and an adaptive KL penalty term. The availability of high-quality simulation environments, such as OpenAI Gym\cite{Palanisamy}, Isaac Gym\cite{Serrano-Munoz}, etc., has greatly accelerated reinforcement learning research. These environments provide a testbed for reinforcement learning algorithms, allowing researchers to quickly iterate and evaluate their ideas. The increasing availability of powerful computing hardware, including GPUs and distributed computing clusters\cite{Liang}, has enabled researchers to train large-scale reinforcement learning models efficiently. In reinforcement learning, a policy function $\pi(s|\mathbf{W}):s \rightarrow a$ is a mapping from states to actions, where $s$ is a state, $a$ is an action and $\mathbf{W}$ is parameters of policy neural network. It specifies the behavior of an agent within an environment. Given a state, the policy function determines which action the agent should take to maximize the cumulative reward over time. The policy function serves as the agent's decision-making mechanism. The essence idea of reinforcement learning generally, is the use of value functions to organize and structure the search for good policies\cite{Sutton}.

Reinforcement learning often faces difficulties in grasping long-term dependencies and managing sparse reward signals. One effective strategy to boost learning efficiency, when feasible, is to incorporate knowledge of classical control. Many OpenAI classical control environments can be solved by classical control theory. Take the Lunar Lander environment, a classic environment that simulates the task of landing a spacecraft on the lunar surface, as an example. This problem can be solved by Proportional-Integral-Differential(PID) control\cite{Knospe}. Classical control theory provides a framework for designing and analyzing systems to achieve desired performance objectives through the use of feedback and advanced functional. Classical controller, such as PID controller, is a mapping $P(s):s \rightarrow a$, calculates action vector from the observation or current state of the agent, which provides a set of ground truth for controlling the agent accomplish its task. Then, deep learning neural network can learn PID control in supervised learning manner. By the objective function measures the Euclidean distance between the output of policy function and PID controller, optimization algorithms attain the optimal policy function parameters $\mathbf{W}$ : 
\begin{equation*}
\mathbf{W} = argmin(MSE(\pi(s | \mathbf{W}),P(s)))
\end{equation*}

Moreover, the nonlinearity of neural network can represent mapping from state to action obtained from classical control approach such as Bellman optimality equation\cite{Bellman,Peng,Dreyfus} or Linear Quadratic Regulator.

\begin{equation*}
\pi(s | \mathbf{W}) \approx P(s) 
\end{equation*}

In summary, classical control if available can be provided as ground truth of actions and help train policy function in supervised learning manner. 

\begin{figure}[h!]
    \centering
    \includegraphics[width=0.8\textwidth]{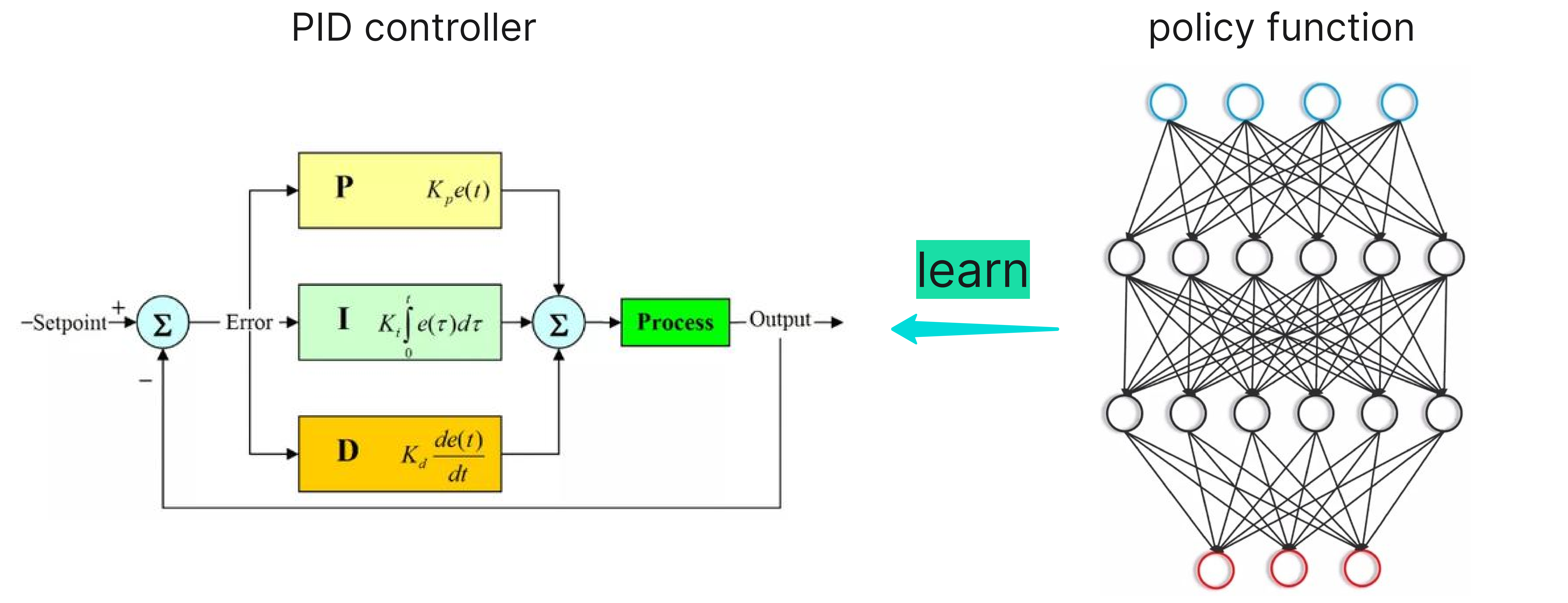} 
    \caption{Available classical control can serve as the ground truth for actions and assist in training the policy function through supervised learning.}
    \label{fig:algorithm}
\end{figure}

\section{Autoencoder Hippocampus Network}\label{sec3}

For different tasks, a same structured policy function but with different sets of parameters values can be trained. Saving and loading corresponding parameters values according to different tasks is a feasible but trivial solution. If we learn a wealth of skills, such as reading, cooking, riding, etc., then save isolated countable files in the brain will make the file system hardly manageable, which is very unlikely the case. Thus, abstract memorization, effective and efficient encoding and decoding procedure, plays a very essential role in learning, inference and execution.

Many neural networks are capable of memorization, including recurrent neural network, Long Short-Term Memory\cite{Sherstinsky}, gate recurrent unit\cite{Gao}, Hopfield network\cite{Hopfield}, autoencoder\cite{Han,Mendoza-Léon} and so forth. Memorization module concerns two key questions what information in what format shall be restored in memory and how to retrieve information from memory. An autoencoder is a type of neural network that is trained to learn efficient data encoding and decoding\cite{Zhai}. It consists of two components: an encoder that compresses the input data into a lower dimensional vector, and a decoder that decompresses the vector back into a representation that is similar to the original input. Autoencoder is more suitable for memorization module over other neural networks because the encoder-decoder architecture of autoencoder make it more convenient and straightforward to retrieve stored information.

In this design, parameters of policy function $\mathbf{W}$ are memorized by autoencoder hippocampus network. Building an autoencoder for parameters of policy function, the encoder maps to learned parameters value to a reduced dimensional latent layer vector called skill vector(or task vector) and the decoder maps latent layer vector, i.e. skill vector, to original parameters tensor. The latent layer of autoencoder provides an interface for retrieving restored information. 


The memorization autoencoder is a mapping consisting of an encoder $E:R^m \rightarrow R^n$ and a decoder $D:R^n \rightarrow R^m$. Policy function's parameters $\mathbf{W}$ memorized by the autoencoder is  
\begin{equation*}
\mathbf{W}^{\prime} = D(E(\mathbf{W}))
\end{equation*}
Ideally, for a well-trained autoencoder, $\mathbf{W}^{\prime}$ should closely resemble $\mathbf{W}$, and the degree of similarity between them can be quantified by the Euclidean distance metric. The essence of this design lies in utilizing the autoencoder hippocampus network to derive a single policy function, parameterized by a set $\{\mathbf{W}_i\}$, that is capable of executing actions tailored to various corresponding tasks.

The autoencoder hippocampus network memorizes the learned parameters tensor. Importantly, memorization process discussed here doesn’t remember anything related to state, action, reward, etc, it only remember the policy function’s optimal parameters. Thus, it is parameters memory only. Given a skill vector, decoder of autoencoder recall (or generate) parameters tensor and assigned them to policy function. The policy function plays a role of capacitor, what parameters tensors are filled in is task oriented.

The skill vector $\mathbf{S}$ is encoded policy function parameters via the encoder: 
\begin{equation*}
\mathbf{S} = E(\mathbf{W}) 
\end{equation*}
The process of decoder recalling parameters $\mathbf{W}$ from the skill vector $\mathbf{S}$ can be written as: 
\begin{equation*}
\mathbf{W} = D(\mathbf{S}) 
\end{equation*}
Then, the action of policy function upon state s reads:
\begin{equation*}
a = \pi(s | \mathbf{W})  = \pi(s | D(S)) = \pi(s | D(E(\mathbf{W}))) 
\end{equation*}

\begin{figure}[h!]
    \centering
    \includegraphics[width=0.9\textwidth]{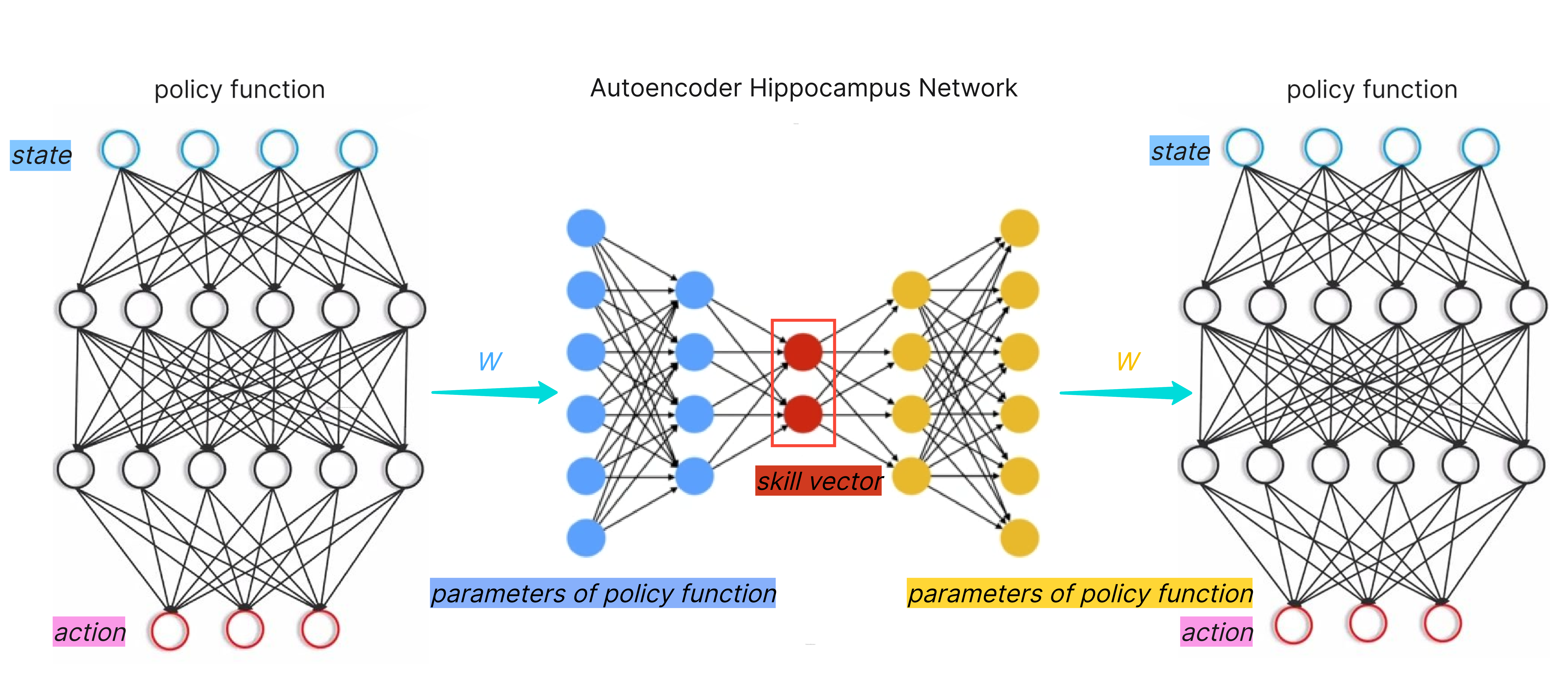} 
    \caption{The parameters of the policy function can be encoded, stored, and retrieved via an autoencoder hippocampus network.}
    \label{fig:algorithm}
\end{figure}

It is impossible in real world to enumerate endless tasks or skills, that is, the learning and execution process shall be extensible and inferable. A person who has learned how to speak does not necessarily know how to ride a bicycle. Conversely, someone who knows how to ride a bicycle is quite likely to possess the skills to ride a motorcycle as well. The skill vectors within the latent layer of an autoencoder embody the concept of Euclidean distance, where a shorter distance signifies a higher degree of skill relevance and, consequently, a greater similarity in parameter values. The memorization module of an autoencoder is capable of inference, including zero-knowledge inference. Provided a skill vector, no matter whether it learned before, as an input for decoder of autoencoder, it will map to or generate a set of corresponding parameters value. Then, for execution phase of a task, policy function loads corresponding parameters value and behave accordingly. 

The design of the memory-embedded policy function involves maintaining a fixed architecture for the policy function while utilizing a memory module to store the parameter values of the policy function during the learning process. During the exploitation procedure, the memory module is used to recall and assign the appropriate parameter values to the policy function. This approach allows the policy function neural network to maintain a consistent architecture but also to possess dynamic parameters that are tailored to specific tasks.

\section{Execution via traversing skill vectors graph}\label{sec4}

The brain has the capability to divide a task to a set of subtasks based on common sense, knowledge, interference and so forth. Some subtasks may have higher priority than the others and form a graph structure in nature. The homeomorphic relationship between subtasks and skill vectors exhibits a one-to-one, bijective mapping. Equivalently, breaking down a task into a graph of subtasks is akin to constructing a graph for skill vectors. Let $G=(\mathbf{S},e)$ be a task graph with subtask vertices $\mathbf{S}$ and edges e. Graph Neural Networks(GNNs)\cite{Zhou,Zhang} capture the dependencies and structures within graphs, making them suitable for tasks such as node classification, link prediction, and graph classification\cite{Yao}. With the autoencoder, encoded skill vector help recall and deploy parameters tensor to policy function, providing an interface for tasks graph management module. When a complicated task can be divided to a topological structure of vector skills set $\{\mathbf{S}_i\}$, the accomplishment of the task can be executed according to skill vectors graph traversal: 
\begin{equation*}
a_i = \pi(s | W_i) = \pi(s | D(\mathbf{S}_i)) = \pi(s| D(\mathbf{S}_i))  
\end{equation*}

\begin{figure}[h!]
    \centering
    \includegraphics[width=0.8\textwidth]{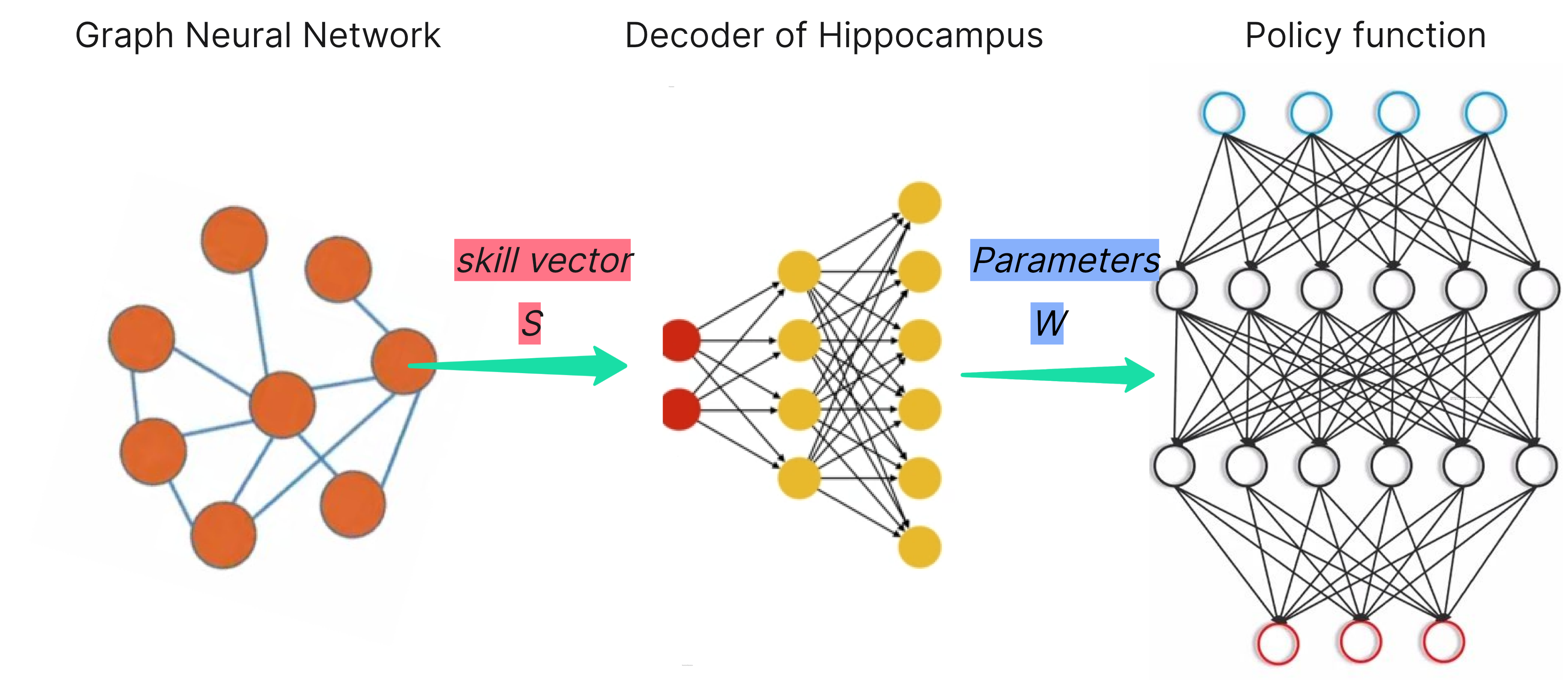} 
    \caption{The latent layer of an autoencoder hippocampus network provides an interface for retrieving the parameters of the policy function. By traversing the graph of skill vectors, a complex task can be completed, and a corresponding sequence of behaviors can be generated.}
    \label{fig:algorithm}
\end{figure}

In summary, the minimal prerequisites for generating a sequence of control encompass: 1) an autoencoder hippocampus mechanism capable of memorizing and recalling parameters of the policy function; 2) construct skill vectors topological graph for the task; and 3) dynamics related to traversing the skill vectors graph.

\section{Conclusion}\label{sec5}

This design is an embedding homogeneous system using a neural network to manage and control another neural network. The main ingredient of the design is to integrate an autoencoder hippocampus network and tasks graph neural network management into the system, dynamically assigning parameters tensor to the policy function by an autoencoder hippocampus network, separating memorization and tasks execution from decision-making, creating a more adaptive and inferable system for different tasks. The multi-functionalities of the deep learning model is a fascinating and promising feature. With the embedding of autoencoder hippocampus network, the system exhibits rich, diverse and complicated dynamical behavior. The accomplishment of a task is akin to a symphony, orchestrating the harmony of motivation, learning, memorization, and execution, with memorization serving as the pivotal nexus connecting all its components.


\section{Data availability}
The data that support the findings of this study are available from the corresponding author upon reasonable request.


\bibliography{sn-bibliography}

\end{document}